\documentclass[10pt,twocolumn,letterpaper]{article}
\usepackage[pagenumbers]{cvpr}
\definecolor{cvprblue}{rgb}{0.21,0.49,0.74}
\usepackage[pagebackref,breaklinks,colorlinks,allcolors=cvprblue]{hyperref}
\usepackage{graphicx}
\usepackage{multirow}
\usepackage{booktabs}
\usepackage[accsupp]{axessibility} 
\usepackage{xcolor}
\usepackage{colortbl}
\usepackage{float}
\usepackage{enumitem}
\usepackage{orcidlink}
\usepackage{stfloats}

\author{
{Xi Wang}\textsuperscript{1} \hspace{.7cm} 
{Yichen Peng}\textsuperscript{2} \hspace{.7cm}
{Heng Fang}\textsuperscript{3} \hspace{.7cm} 
{Yilin Wang}\textsuperscript{4}\\
{Haoran Xie}\textsuperscript{5} \hspace{.7cm} 
{Xi Yang}\textsuperscript{1}\thanks{Corresponding author} \hspace{.7cm}
{Chuntao Li}\textsuperscript{1}
\\
\textsuperscript{1}Jilin University.\\
\textsuperscript{2}Tokyo Institute of Technology.\\
\textsuperscript{3}KTH Royal Institute of Technology.\\
\textsuperscript{4}Adobe Research\\
\textsuperscript{5}Japan Advanced Institute of Science and Technology (JAIST).
}
\begin{document}
\title{FilterPrompt: A Simple yet Efficient Approach to Guide Image Appearance Transfer in Diffusion Models}

\twocolumn[{
\renewcommand\twocolumn[1][]{#1}
\maketitle
\begin{center}
    \vspace{-6mm}
    \captionsetup{type=figure}
    \includegraphics[width=\textwidth]{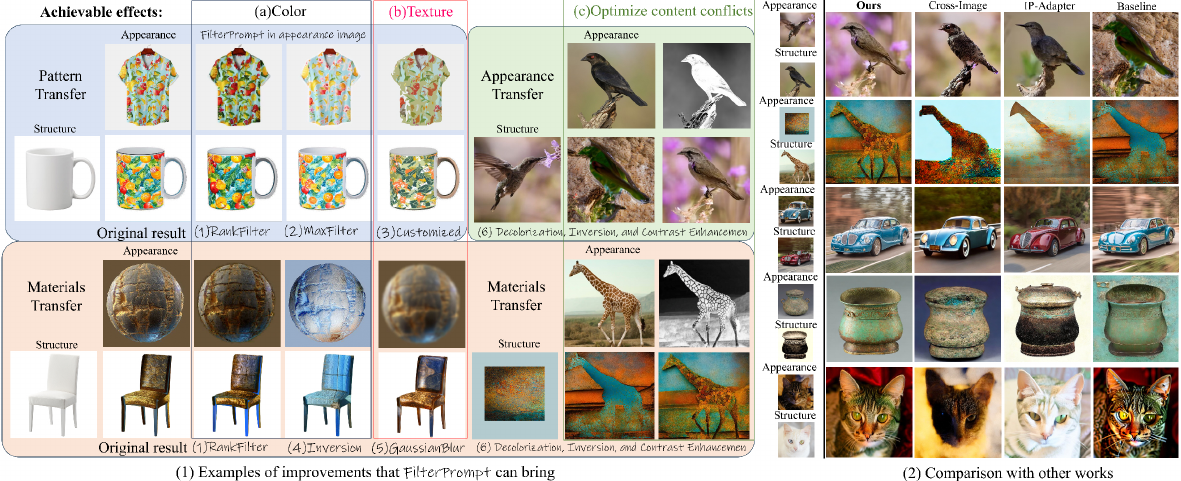}
    \captionof{figure}{
    \textbf{The comparison of generated results.} Our approach FilterPrompt enables appearance transfer in multiple domains at local, object-centric, and full-graph levels. 
    Compared to previous works like Cross-Image~\cite{Cross-imageAttention}, IP-Adapter~\cite{Ip-adapter} and baseline (IP-adapter~\cite{Ip-adapter}  +ControlNet~\cite{ControlNet}), our approach can help the model better preserve the geometric properties of structural images while maintaining consistent color distribution and texture features with appearance images. 
    }
    \label{fig:1_teaser}
\end{center}
}]
\begin{abstract}
In controllable generation tasks, flexibly manipulating the generated images to attain a desired appearance or structure based on a single input image cue remains a critical and longstanding challenge. 
Achieving this requires the effective decoupling of key attributes within the input image data to achieve representations accurately.
Previous works have concentrated predominantly on disentangling image attributes within feature space. However, the complex distribution present in real-world data often makes the application of such decoupling algorithms to other datasets challenging. Moreover, the granularity of control over feature encoding frequently fails to meet specific task requirements.
Upon scrutinizing the characteristics of various generative models, we have observed that the input sensitivity and dynamic evolution properties of the diffusion model can be effectively fused with the explicit decomposition operation in pixel space. This allows the operation that we design and use in pixel space to achieve the desired control effect on the specific representation in the generated results. 
Therefore, we propose FilterPrompt, an approach to enhance the effect of controllable generation. It can be universally applied to any diffusion model, allowing users to adjust the representation of specific image features in accordance with task requirements, thereby facilitating more precise and controllable generation outcomes.
In particular, our designed experiments demonstrate that the FilterPrompt optimizes feature correlation, mitigates content conflicts during the generation process, and enhances the effect of controllable generation, as shown in Figure~\ref{fig:1_teaser}.

\end{abstract}

\section{Introduction}
In controllable image generation, achieving flexible control over the appearance attributes of objects in the generated images, such as texture and material, remains a research focus~\cite{bengio2013representation}.
Some researchers concentrate on refining the data, aiming to acquire the low-dimensional feature representations of input images~\cite{DiffusionAutoencoders, CVPR2022_imageD, wu2023uncovering}. 
Concurrently, another faction of researchers is interested in improving the model architecture. They employ deep learning techniques such as autoencoders (AE), variational autoencoders (VAE), and generative adversarial networks (GAN) to fine-tune feature extraction and processing methods, thereby enhancing the capacity of models to handle complex data autonomously~\cite{2016betaVAE, 2016VAE-GAN, 2017wasserstein, 2018DCGAN}.

Specifically, controllable generation typically follows two ways:
\textit{First}, disentangling the characteristics of an input image in the feature space and obtaining feature representations relevant to the control objective. 
Subsequently, the network regulates the degree to which these feature representations are expressed in the generated image, employing a diverse array of meticulously designed loss metrics~\cite{NIPS2016_InfoGAN, NEURIPS2020_ICAM, CVPR2021_PISE, CVPR2022_imageD, ICCV2023_StyleDiffusion, 2023Pairdiffusion}.
\textit{Second}, involving incorporating a conditioning mechanism into the architecture of the model. These works improve the capacity of the model to integrate control conditions while learning the target domain's data distribution~\cite{CGAN, NEURIPS2020_variational, CVPR2023_SpecialistDiffusion, SIGG2023sketch}. 
Then, the generated images exhibit an artistic effect similar to the appearance of the training data set and consistent with the input image structure.

However, the above controllable generation ideas have their limitations.
On the one hand, mapping different data domains to the same feature space will incur high computational costs.
Information loss in this process cannot be avoided while there is also the problem of attribute entanglement between the representations obtained in the feature space~\cite{CVPR2021_causalvae, brehmer2022weakly}. 
On the other hand, using machine learning algorithms to train style mappings may not have the same level of interpretability as traditional mathematical modeling approaches, and the training of such models often requires expensive data collection~\cite{2023instructpix2pix, LCLR2022_Learning}.

\begin{figure}
    \centering
    \includegraphics[width=1\linewidth]{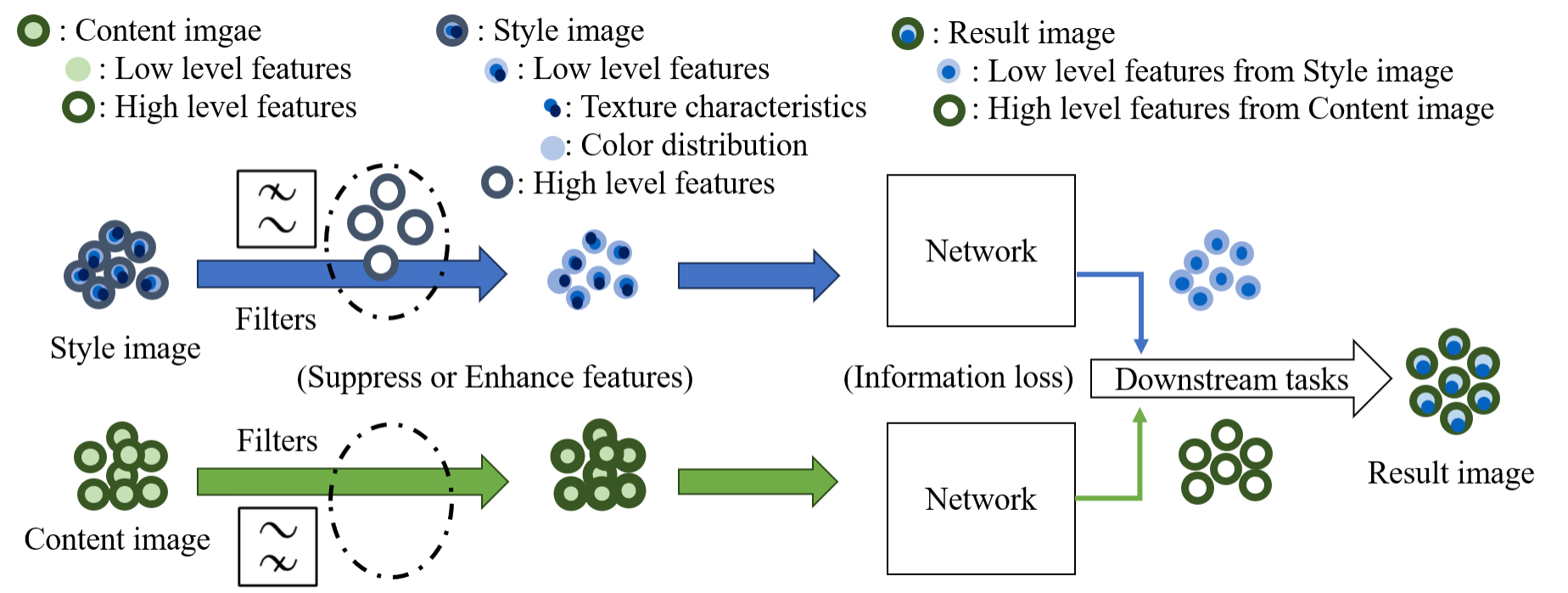}
    \caption{\textbf{Our FilterPrompt.} When diffusion models extract image features, strategically incorporating filtering operations enables targeted suppression or enhancement of particular feature distributions. The filters enhance the performance of diffusion models to improve the quality of generated images. }
    \vspace{-6mm}
    \label{fig:2_2}
\end{figure}

Hence, unlike previous endeavors that focused on refining algorithms and models within the feature space, we redirect our focus toward the pixel space. 
Intuitively, certain semantic features, often indistinguishable in models, exhibit discernible distribution discrepancies visible to the naked eye in the pixel space. 
Following experimental comparisons of various mainstream generative models, we observe that the diffusion model possesses properties of input sensitivity and dynamic evolution which are very suitable for image processing operations in the pixel space. 
This integration enables image processing operations performed in the pixel space for a specific feature distribution of the input image and can achieve the desired control effect in the generated results. 
Therefore, we name this approach FilterPrompt as shown in Figure~\ref{fig:2_2}.

To validate the aforementioned statement, we build a framework for experimental testing based on the existing pre-trained model and use it to demonstrate the impact of FilterPrompt on the control effect of appearance features and structure features of the generated image. 
Then, we conduct quantitative analyses of the generated results. 
These analyses demonstrate that the FilterPrompt optimizes feature correlation, mitigates content conflicts during the generation process, and enhances the model's control capability.
At the same time, our approach can be easily generalized to more complex combinations without requiring additional training and is highly interpretable.

In summary, our contributions are as follows: 
1. We introduce a new approach, FilterPrompt, aimed at optimizing model control effects. Our approach can be customized for various input data based on specific task requirements and combined with the diffusion model to achieve the expected control effect. 
2. We analyze how FilterPrompt facilitates the desired control effect within the diffusion model framework. Additionally, we have designed experiments to provide explanations to demonstrate the feasibility of our approach.
3. Our experiments encompass a range of tasks for the local, object-centered, and full-graph appearance transfer. We present the application of FilterPrompt in these tasks and compare its performance with various other models. Through experimentation, we substantiate the efficacy of our approach in image transfer.

\section{Related Work}
\subsection{Controllable Generation in Diffusion Modeling}
The research on the controllable image generation task in the diffusion model can be broadly divided into three stages.

The \textit{first stage} is mainly based on the iterative denoising process and achieves controllable generation effects by rationally using the input image to generate a deterministic guided generation paradigm~\cite{choi2021ilvr, 2021sdedit, 2022repaint}.
This stage of the work controls the semantic similarity between the generated image and the input image by influencing the proportion of information mixing in the denoising network of U-Net.
The \textit{second stage} is based on the image generation strategy guided by the display classifier.
Optimization is performed by adding gradient information from the classifier to the loss function.
This idea first originated from classifier guidance\cite{song2020score} and further advanced in\cite{NEURIPS2021, rombach2022high}.
Since then, numerous studies have broadened the scope of classifiers, extending the classification guidance of the diffusion model to encompass diverse modalities such as text, images, and other multi-modal data~\cite{2022blendeddiffusion,2022diffusionclip,2023uncovering,2023Pairdiffusion,2023more}.
The \textit{third stage} marks the era of large models based on implicit classifiers.
To address the issue of declining diversity in classifier guidance, classifier-free guidance strategy~\cite{2022classifierFreeGuidance} emerges later. 
This approach involves decomposing the gradient guidance from the explicit classifier into two components. 
One component is an unconditionally generated gradient prediction model, akin to the conventional DDPM~\cite{2020DDPM}. The other is a gradient prediction model based on conditional generation, conceptualized as a U-Net network with an overlay of a cross-attention mechanism.
The success of this approach has catalyzed the evolution of a variety of subsequent image editing technologies. These technologies utilize the diffusion model as a foundational framework and integrate the attention mechanism, resulting in notable progress in the application of the diffusion model across diverse fields. Notable projects in this domain include DALLE-2~\cite{2022DALLE-2}, DreamFusion~\cite{2022_Dreamfusion}, Stable Diffusion Model (SDM)~\cite{2022SDM}, and more.

Here, our primary focus lies on Grounded Generation and Layout-driven Generation within the context of the controllable generation problem. Representative works in this area include GLIGEN~\cite{2023gligen}, ControlNet~\cite{ControlNet}, and IP-Adapter~\cite{Ip-adapter}. We aim to investigate the generative capabilities of diffusion models in addressing semantic-level conditional guidance, particularly in scenarios with limited sample sizes. For example, appearance transfer task~\cite{suri2024grit}. It needs to preserve the structure of the target image while applying the desired appearance attributes. 

\subsection{Explicit Decomposition }
Explicit decomposition is aimed at breaking down the representation of data or a model into simpler, more independent components or factors. Specifically, this process involves splitting a high-dimensional representation space into multiple low-dimensional subspaces, each responsible for encoding a specific aspect or attribute of the data. 
Through explicit decomposition, neural networks can more easily understand various aspects of the data, such as geometry, color distribution, texture, etc., in the image~\cite{huang2023composer}.

Traditionally, filtering algorithms have been considered the explicit decomposition approach as they can break down input data into components at different frequencies or spatial scales.  
Examples of common filters include the Gaussian filter, Sobel filter~\cite{1968sobel}, Adaptive filter~\cite{1995adaptiveFilter}, and Gabor filter~\cite{1946gabor, 2012gabor_texture}. 
These filters can weigh data at different frequencies or scales to suppress or enhance specific features. 
Therefore, performing preprocessing operations may help neural networks better understand the structure and features of data in order to acquire more refined representations.
\section{Prerequisite}
\begin{figure*}[t]
    \centering
    \vspace{-6mm}
    \includegraphics[width=1\textwidth]{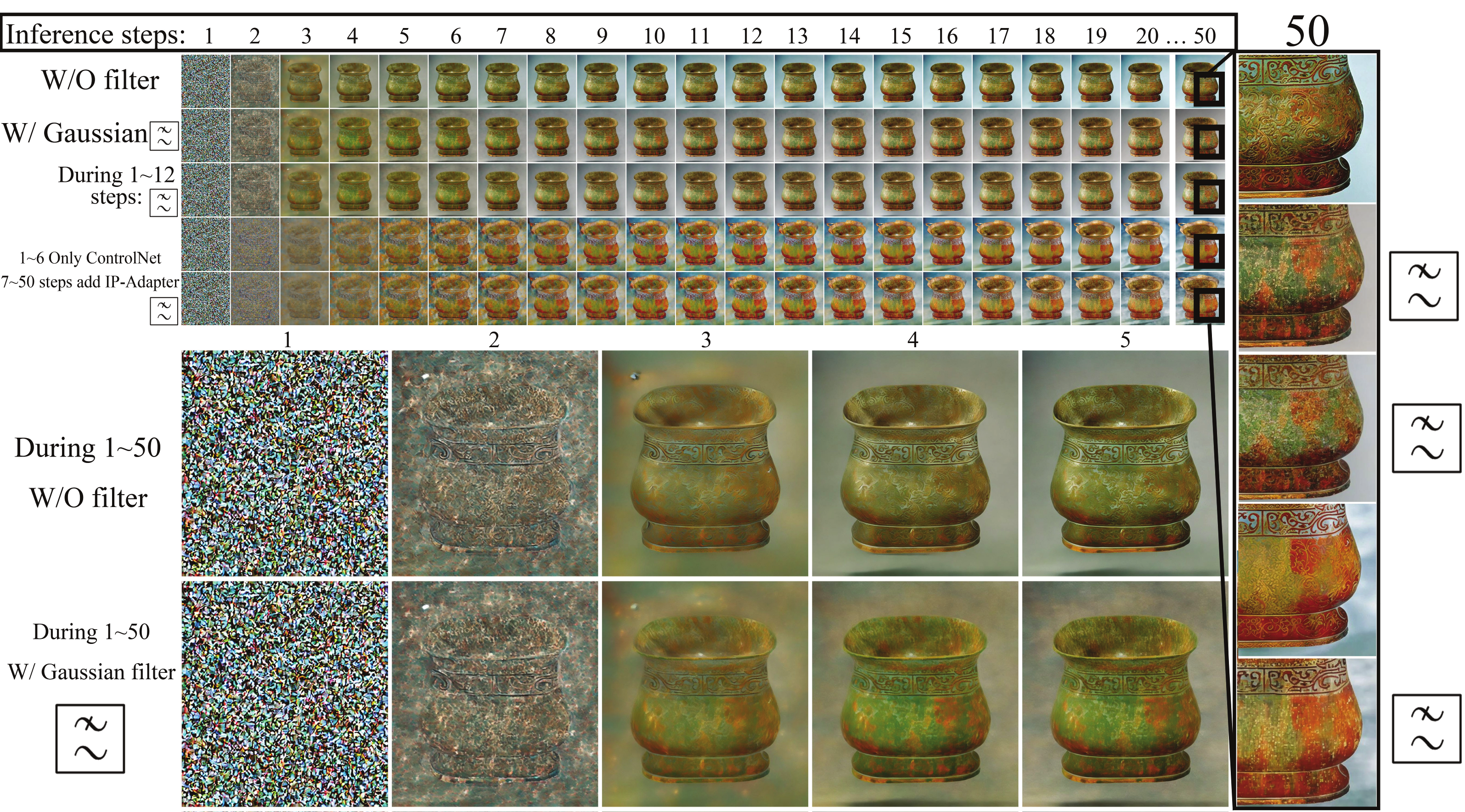}
    \caption{\textbf{Filter impact on sampling inference process.} After applying a Gaussian filter, the underlying texture in the sampled images changes from a distribution resembling arc patterns to a point-like distribution. Additionally, as shown in the enlarged illustration on the right, it is evident that the use of filters consistently disrupts the expression of redundant pattern features.
    }
    \label{fig:4_3}
    \vspace{-6mm}
\end{figure*}
\subsection{Prompt Impact on Diffusion Models}
The forward diffusion defines a known Gaussian translation process. Then the image intermediate quantity $x_{t}$ at each moment of the forward process can establish a unique relationship with the input image $x_{0}$ as:

\begin{equation}
    \label{eq:forward_X0_Xt}
    q(x_t|x_0):=N(x_t;\sqrt{\bar{\alpha_t}}x_0,(1-\bar{\alpha_t})I),
    \;\;\;
    x_t=\sqrt{\bar{\alpha_t}}x_0+(1-\bar{\alpha_t})\epsilon
\end{equation}
The reverse denoising process of the diffusion model can be seen as a migration process to the target data distribution.
During this process, the model will continuously try to reduce noise and be guided by the condition $c$ to restore the structure or characteristics of the specified data. Every migration process can be expressed as:
\begin{equation}
    \label{eq:reverse_sum}
    x_{t-1}=\frac{1}{\sqrt{\alpha_{t}}}\left(x_{t}-\frac{1-\alpha_{t}}{\sqrt{1-\bar{\alpha}_{t}}}\epsilon_{\theta}\left(x_{t},c,t\right)\right)+\sigma_{t}z
\end{equation}

\begin{figure*}[t]
    \centering
    \vspace{-6mm}
    \includegraphics[width=\textwidth]{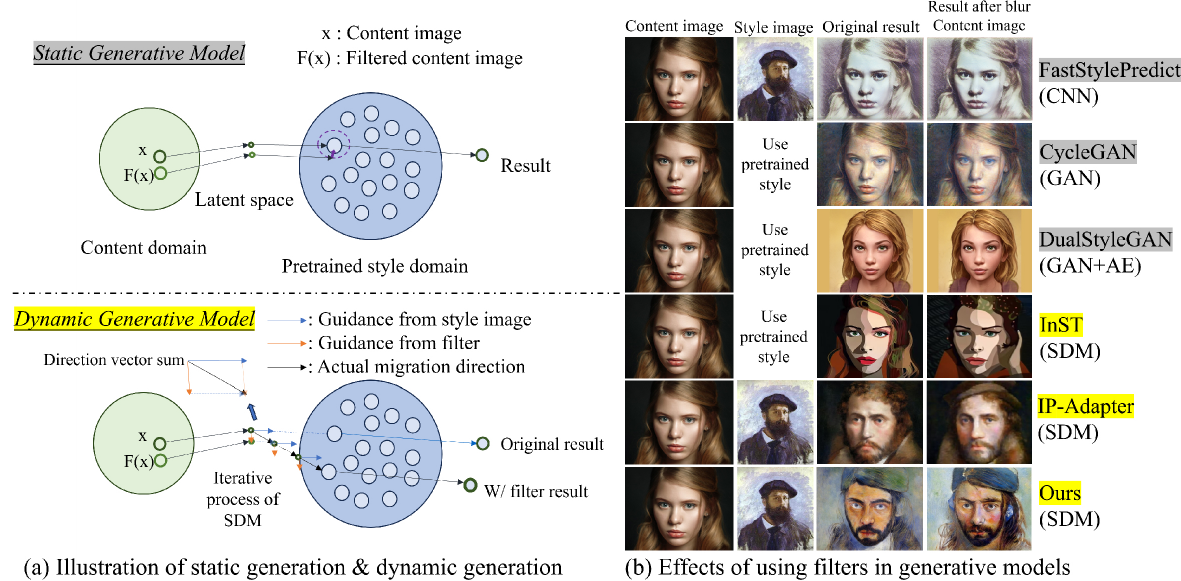}
    \caption{(a) shows the illustration of filter's impact on the corresponding sampling inference stages in the static generative model and the dynamic generative model.
    (b) gives a comparison of the results obtained by applying filter to some works of generative models. The gray background represents traditional work based on GAN and AE architectures~\cite{FastStylePredict, 2017_CycleGAN, CVPR2022_DualStyleGAN}. The yellow background represents work based on Diffusion~\cite{INST, Ip-adapter}. Comparing the results, we can intuitively see that filter operation has a more significant impact on diffusion models. 
    }
    \vspace{-6mm}
    \label{fig:3_1}
\end{figure*}

The focus of this process is the network’s prediction of noise distribution: $\epsilon_{\theta}\left(x_{t},c,t\right)$.
This prediction is affected by the current moment $x_{t}$ and condition $c$, and condition $c$ comes from the external reference image $y$ mapping result. 
Therefore, if we perform filters $f_{\gamma}$ on either side of the input like making $c=\mathrm{Encoder}(y)$ become ${c{'}=~\mathrm{Encoder}(f_{\gamma}(y))~}$, it will affect the prediction results of the noise distribution $\epsilon_{\theta}$, and even affect the migration direction of the generated distribution at that time node as illustrated in Figure \ref{fig:3_1}. 
The impact caused by conditions $c$ or the input structure image $x_{0}$ would be reiterated at each sampling instance, and the minor changes introduced by filters would be involved in the entire generation process. 

We observe that after applying filtering operations to a specific feature distribution in pixel space, the degree of expression for that distribution aligns with expectations.
Moreover, the filtering operations targeting a specific feature distribution do not affect the expression of other features. 
This indicates that the influence of filters is independent in the feature space from the encoding of other representations. 
Therefore, prompts in pixel space offer a lightweight and convenient way to optimize the entanglement between feature representations in the diffusion model.

Based on the framework of combining ControlNet~\cite{ControlNet} and IP-Adapter~\cite{Ip-adapter}, we examine the impact of filters at various stages of the sampling inference process and elucidate that the filter plays a guiding role in guiding the Gaussian distribution of the current data toward the target distribution during the migration and diffusion process.

We perform a detailed analysis of the sampling inference stage for the task of converting a bronze sketch to a photo, as shown in Figure \ref{fig:4_3}. The sampled results vividly illustrate that detailed representations of arc patterns initially present in the early stages are weakened by the Gaussian filter, manifesting as point-like distributions. Besides, the Gaussian filter disrupts the continuous semantic expression of patterns. Concurrently, subsequent texture generation doesn't emerge the negative impact of full-graph blurring, so it illustrates the property that the filtering operation is only effective for specific feature distributions. 
 
The third row of the evolution sequence in Figure \ref{fig:4_3} illustrates the impact of applying the Gaussian filter only in the first 12 steps, where we observe the absence of redundant pattern features in the final generation result.
A set of comparisons in lines 4 and 5 also showcase the effectiveness. In the first 6 steps, only ControlNet is utilized to regulate the structural layout. 
Introducing IP-adapter in the 7th step guides appearance features. The final results reveal that, even if the Gaussian filter does not initially suppress redundant features, it remains effective in later stages during the generation of detailed textures. 
The above findings highlight that the impact of filtering on the diffusion process is intuitive, controllable, and predictable.

\subsection{Static Generation vs. Dynamic Generation}
We pay attention to the fact that in the sampling inference stage, the diffusion model has different dynamic evolution properties from the traditional generative model. 
In the basic theory of the diffusion model, the process of generating an image starts from a simple noise image~\cite{2020DDPM}, and through multiple iterations, the noise in the image is gradually removed until a final clear image is generated. 
From this perspective, the diffusion model generation process is continuous in time, and the image is generated through a gradually changing process. 
Therefore, we believe that the gradual evolution of the diffusion model makes it a dynamic generative model.

In the early days, the fast style transfer algorithm~\cite{gatys2016image, li2016combining, FastStylePredict} is based on CNN, AE, and GAN~\cite{choi2020stargan, 2017_CycleGAN, CVPR2022_DualStyleGAN}, and their fitting effect on the target data distribution depends on the design of the training process. 
During the training process, the model attempts to capture the statistical characteristics of the entire data distribution and obtain the maximum likelihood representation of the data distribution. 
When training is completed, the generated samples exist statically in the latent space. Therefore, we consider this type of model to be the static generative model.

We noticed that prompts in pixel space may exhibit certain limitations in previous static generation models, but it can be well combined with dynamic generation models. 
As shown in Figure \ref{fig:3_1}, we applied the same level of blur interference to the input data of various generative models. The results indicate that the impact of blur prompt on static generative models is relatively weak, while the outcomes of diffusion models show significant changes.
We believe that for static generation models, using prompt on input data during the sampling inference stage will only have a slight impact on the mapping position of this input in the static distribution. And this position is close to the mapping result of the original image $x$. 
Therefore, prompts in pixel space will not significantly affect the generation results of static generation models. 
However, for diffusion models, as indicated by Equation \ref{eq:forward_X0_Xt}, there is a predefined long-term dependency between the current time node image $x_{t}$ and the input image $x_{0}$, in the forward process. 
Therefore, the impact of using prompts on the input structure image $x_{0}$ will be executed again at each sampling, and the minor changes caused by prompts will participate in the entire generation process, thereby exerting a more significant influence on the generation results.
\section{Our Approach}
\begin{figure*}[t]
    \centering
    \vspace{-6mm}
    \includegraphics[width=1\linewidth]{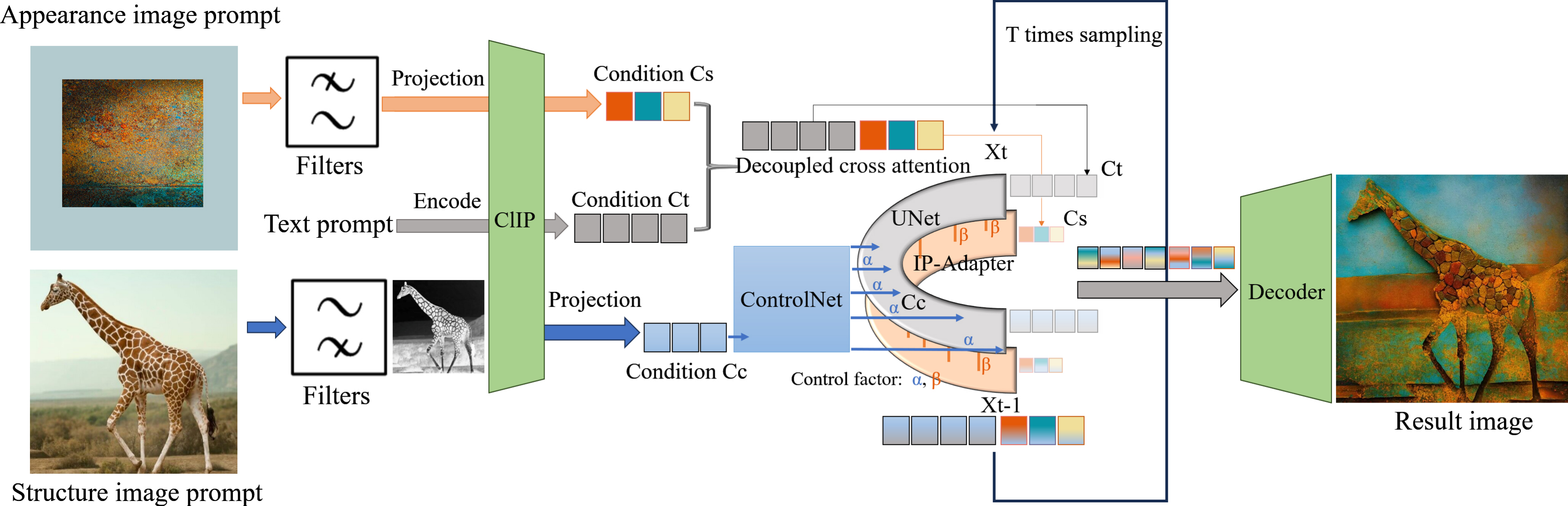}
    \caption{\textbf{Our framework.} The experiment uses ControlNet and IP-Adapter as the baseline and adds combined filtering operations as the expansion. We mapped low-level features in appearance images to global embeddings as $Cs$, concatenating them with SDM default text prompt embeddings $Ct$. The denoising generation processes these parts separately. A segment is managed by ControlNet, projecting latent distributions into a fused distribution controlled by high-level features that is $Cc$. The other part uses IP-Adapter for decoding and guiding low-level feature generation. Intermediate hidden state $x_{t-1}$ from both processes are weighted and summed every sampling time.
    }
    \label{fig:4_1}
    \vspace{-6mm}
\end{figure*}
We propose the adoption of pixel-space methodology, FilterPrompt, to directly manipulate the frequency or distribution characteristics of specific image attributes, thereby influencing the subsequent expression levels of the representation. Our approach offers an intuitive and straightforward approach, and significantly saves computational overhead. 
\subsection{FilterPrompt}
\label{4_1}
We leverage IP-Adapter\cite{Ip-adapter} to obtain conditional encoding $C_{s}$, which controls rendering attributes in the generated image from the appearance reference $x_{app}$.
Concurrently, we employ ControlNet\cite{ControlNet} to obtain the conditional encoding $C_{c}$, responsible for regulating the geometric attributes from the structure reference image $x_{struct}$. 
These encodings, $C_{s}$ and $C_{c}$, represent features from their respective reference images:
\begin{equation}
\begin{aligned}
    C_{s}=\left\{feat_{\{m\}(x_{\{app\}})}\right\}_{\{m=1\}}^{\{M\}}\\
    C_{c}=\left\{feat_{\{n\}(x_{\{struct\}})}\right\}_{\{n=1\}}^{\{N\}}
\end{aligned}
\end{equation} 
However, since $C_{s}$ and $C_{c}$ features are not fully disentangled, using them directly in latent space may cause structural conflicts. Drawing upon the successful performance of prompts demonstrated in the dynamic generation model above, we try to alleviate these conflicts by leveraging FilterPrompt, which adjusts feature frequencies or distributions. For instance, a color-removing prompt $F$ removes color features $feat$, modifying the encodings as follows:
\begin{equation}
\begin{aligned}
      C_{s}^{\prime}=\left\{feat_{\{m\}(F(x_{\{app\}}))}\right\}_{\{m=1\}}^{(M-k)}\quad\\
      C_{c}^{\prime}=\left\{feat_{\{n\}(F(x_{(struct)}))}\right\}_{\{n=1\}}^{\{N-k\}}
\end{aligned}
\end{equation}
The pixel range affected by $F$ is intuitive and convenient, so we can estimate the impact of it on the migration of Gaussian distribution in Equation \ref{eq:reverse_sum} and change the iterative process into:
\begin{equation}
    x_{t-1}=\frac{1}{\sqrt{\alpha_{t}}}\left(x_{t}-\frac{1-\alpha_{t}}{\sqrt{1-\bar{\alpha}_{t}}}\epsilon_{\theta}\left(x_{t},c_{\{s\}}^{\prime},c_{\{c\}}^{\prime},t\right)\right)+\sigma_{t}z
\end{equation}
In conclusion, we advocate for the adoption of our innovative approach, FilterPrompt, which directly manipulates the frequency or distribution characteristics of specific image attributes, thereby influencing the subsequent expression levels of the representation. 
\subsection{Architecture Details}
\label{4_2} 

Our framework are built based on combined filtering operations, ControlNet~\cite{ControlNet} and IP-Adapter~\cite{Ip-adapter}, as shown in Figure \ref{fig:4_1}. 
Subsequently, we map the low-level features in the appearance image to a global embedding $C_{s}$ and concatenate it with the default text prompt embedding $C_{t}$ of SDM. This process can be described as $X_{t} = C_{t}  \bigoplus  C_{s}$.
These two parts in hidden state $x_t$ are processed separately at each denoising generation.
A portion of $x_t$ is delegated to ControlNet, which projects the latent distribution into a fused distribution controlled by high-level features $C_{c}$. The global embedding of another part in $x_t$ utilizes the IP-Adapter for decoding, unfolding, and guiding the generation of low-level features.
We use $x_{t-1}$ to represent the hidden state predicted at the next moment in Equation \ref{eq:reverse_sum}. The intermediate hidden states obtained from both processes are weighted and summed according to Equation \ref{eq:sum_ControlNetAndIP}, achieving the effect of unifying representations related to Structure and Appearance into the latent space of SDM as shown in Equation \ref{eq:sum_ControlNetAndIP}. 
\begin{equation}
\label{eq:sum_ControlNetAndIP}
 X_{t-1} = \alpha \cdot ControlNet +\beta \cdot \lambda \cdot IP-Adapter
\end{equation}
where $\alpha$, $\beta$ are weight control factors and $\lambda$ is scale control factor. 
\begin{figure*}[t]
    \centering
    \vspace{-6mm}
    \includegraphics[width=1\textwidth]{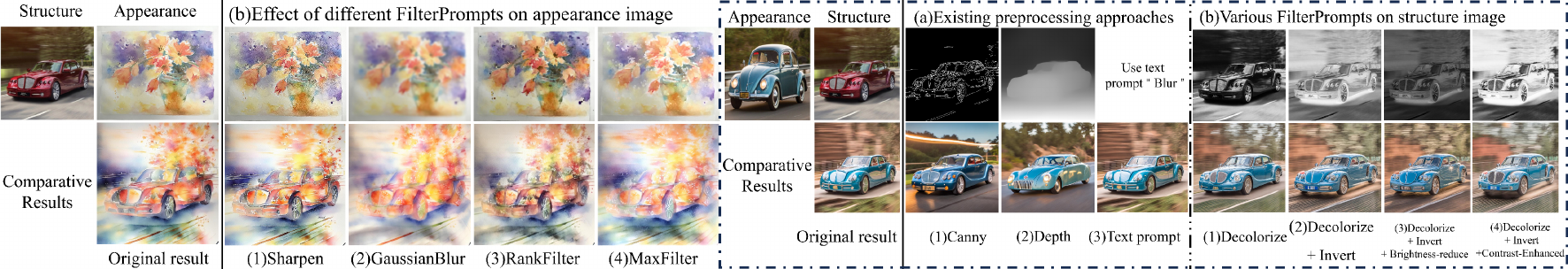}
    \caption{
    \textit{Figure-left}: The effect of FilterPrompt applied to the appearance image. Among the results, the Sharpen filter enhances the expression of fine strokes, while the Gaussian filter blurs detailed stroke information. This demonstrates that FilterPrompt can significantly influence appearance information, aligning with our expectations.
    \textit{Figure-right}: The effect of FilterPrompt applied to the structure image. The generated results show that the $FilterPrompt_{struct}$ in FilterPrompt(4) best preserve structural information. Specifically, these filters allow high-fidelity reproduction of critical vehicular details, such as the exhaust window and headlights, which existing preprocessing methods fail to replicate.
    }
    \label{fig:4_filterPromptEffect}
\end{figure*}
\begin{figure*}[t]
    \centering
    \vspace{-3mm}
    \includegraphics[width=1\textwidth]{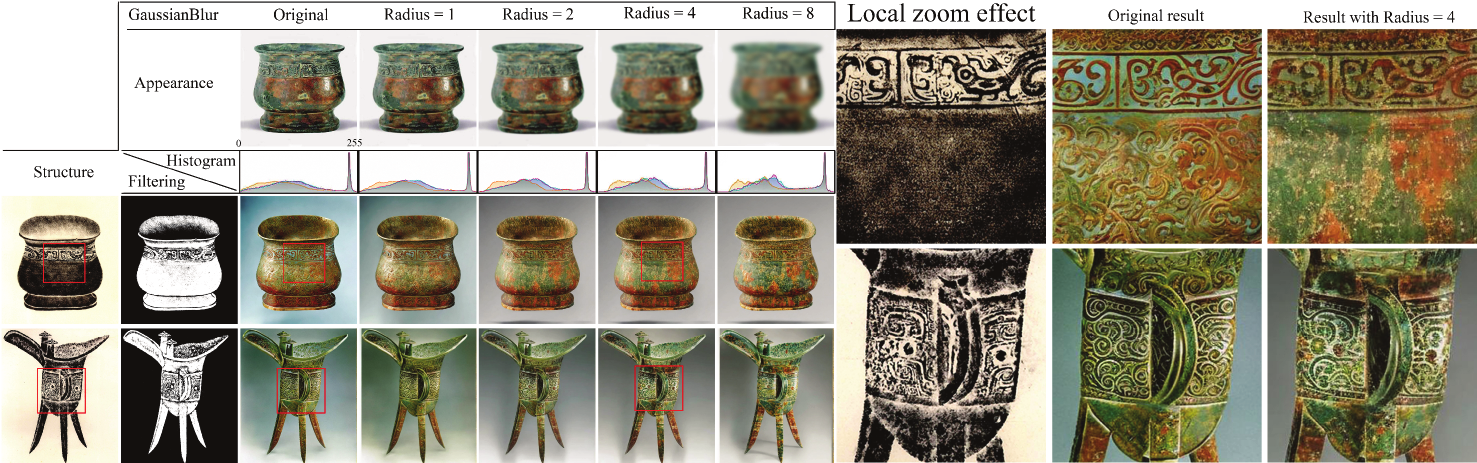}
    \caption{\small\textbf{Impact of different kernel sizes in FilterPrompt on the generated results.} 
    In this example, we utilize $FilterPrompt_{struct}$ mentioned before on the structure image. Simultaneously, a Gaussian filter is applied to the appearance image. 
    The outcomes highlight the effectiveness of $FilterPrompt_{struct}$ on the structure image in preserving the geometric attribute information of the bronze. Additionally, increasing the Gaussian kernel size helps reduce the representation of redundant pattern information in the appearance image, thus addressing content conflicts in the generated results.
    }
    \vspace{-3mm}
    \label{fig:4_kernelSize}
\end{figure*}
\subsection{Effects of Various FilterPrompts}
\label{4_3}
We define structure as the geometric features in the structure image and appearance as the rendering features from the reference image’s color and texture. We then assess baseline performance on these features before and after applying FilterPrompt (see Figure \ref{fig:4_filterPromptEffect}).

Firstly, we apply our approach on the ControlNet path for controlling structural details.
The comparison of results among existing image preprocessing approaches and FilterPrompt indicates that our approach retains more details in the generated results and brings the colors closer to the target appearance. 
The combination of operations (including the ITV-R 601-2 luma transform method for decolorization, inversion, and contrast enhancement of the grayscale image) is succinctly referred to as $FilterPrompt_{struct}$. As illustrated in Figure \ref{fig:4_filterPromptEffect}-FilterPrompt(4), $FilterPrompt_{struct}$ best preserves structural information by enhancing brightness for clearer outlines, demonstrating FilterPrompt's clarity and interpretability.

We then analyze the IP-adapter path, responsible for appearance, before and after applying FilterPrompt. Figure \ref{fig:4_filterPromptEffect} shows that applying noise-processing filters to the appearance image affects the generated stroke details, aligning with the filter’s effects—e.g., Sharpen enhances fine strokes, while Gaussian blurs them. This verifies FilterPrompt’s capacity to control appearance.

Taking a step further, we explore the control effect of FilterPrompt on both paths in the baseline. 
In appearance task as shown in Figure \ref{fig:4_kernelSize}), we aim is to transfer a bronze sketch to a photo with a specified appearance image without altering geometric features from the structure image. For structure control, $FilterPrompt_{struct}$ in the ControlNet path was used, but initial results showed redundant patterns. Applying a Gaussian filter in the IP-Adapter path suppressed high-frequency noise. When the Gaussian kernel increased to 4, redundant features diminished significantly.

\section{Experimental Results}
\begin{table*}
\centering
\vspace{-6mm}
\captionof{table}{\textbf{Metrics evaluation.} The results demonstrate that FilterPrompt achieves better performance in preserving structure, shape, and edge similarity, as well as in maintaining feature distribution similarity, texture differences, image quality, and color histogram correlation. 
We highlight the best value in \textcolor{black}{\colorbox[HTML]{FFCCC9}{red}}, and thesecond-best value in \textcolor{black}{\colorbox[HTML]{FFFFC7}{yellow}}.}
\scalebox{0.95}{
\begin{tabular}{ccccccc}
\hline
 & Structure Preservation & Shape and Edge & Feature Distribution & Texture & Quality & Color Correlation \\ \cline{2-7} 
\multirow{-2}{*}{} & \textbf{SP↑} & \textbf{CD↓} & \textbf{FID↓} & \textbf{GLCM↓} & \textbf{PSNR↑} & \textbf{CHC↑} \\ \hline
\textbf{Cross-Image} & 0.7791 & 5.4133 & 245.0973 & \cellcolor[HTML]{FFFFC7}0.1376 & 9.4278 & 0.9357 \\ \hline
\textbf{IP-Adapter} & 0.8313 & 4.0967 & \cellcolor[HTML]{FFCCC9}\textbf{210.2189} & 0.1619 & 9.5546 & 0.8004 \\ \hline
\textbf{Baseline} & \cellcolor[HTML]{FFFFC7}0.8547 & \cellcolor[HTML]{FFFFC7}3.3027 & 222.8576 & 0.1618 & \cellcolor[HTML]{FFFFC7}10.5011 & \cellcolor[HTML]{FFFFC7}0.9364 \\ \hline
\textbf{Ours} & \cellcolor[HTML]{FFCCC9}\textbf{0.8799} & \cellcolor[HTML]{FFCCC9}\textbf{2.8092} & \cellcolor[HTML]{FFFFC7}215.8267 & \cellcolor[HTML]{FFCCC9}\textbf{0.1072} & \cellcolor[HTML]{FFCCC9}\textbf{10.5594} & \cellcolor[HTML]{FFCCC9}\textbf{0.9405} \\ \hline
\end{tabular}
}
\vspace{-3mm}
\label{tab:Quantitative_structure}
\end{table*}
\begin{figure*}
    \centering
    \includegraphics[width=\textwidth]{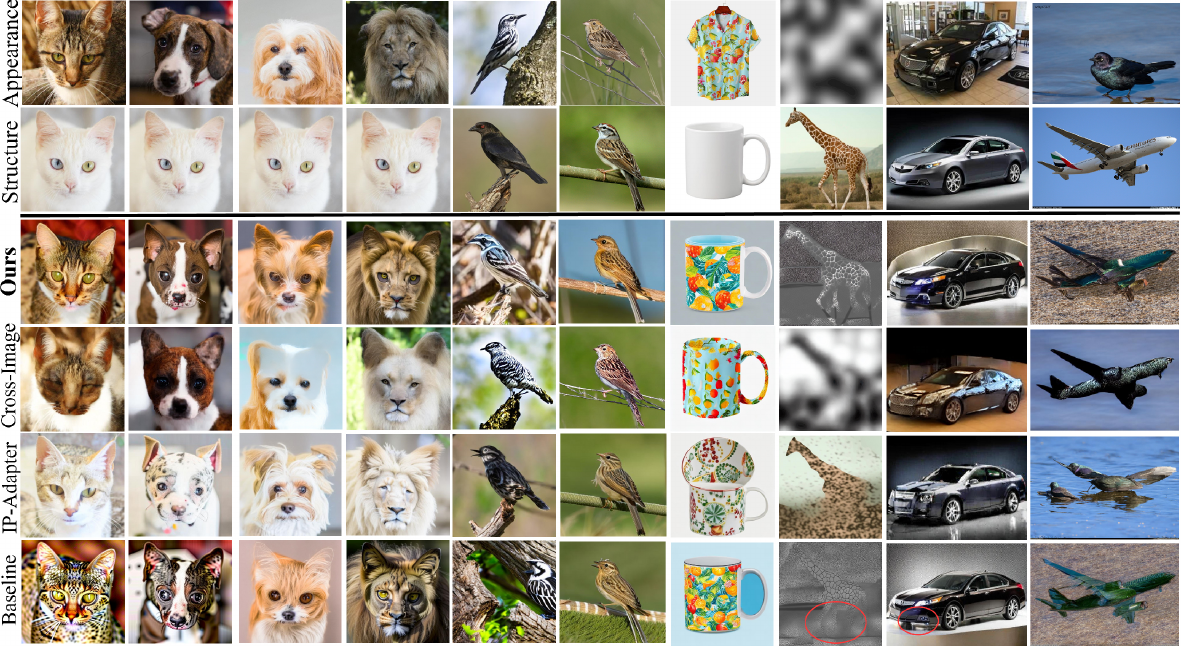}
    \caption{\small\textbf{Comparison with other works.} As shown in the 1st, 2nd, 3rd, 5th, and 8th columns of the figure, the baseline has content conflict problems, the Cross-image attention generates blurry results, and the IP-Adapter cannot accurately transfer the color of the appearance image. Applying $FilterPrompt_{struct}$ on the baseline results can enhance the protection of structural attributes while alleviating content conflicts.
    }
    \label{fig:5_1}
    \vspace{-6mm}
\end{figure*}
\begin{figure*}[t]
    \centering
    \includegraphics[width=0.85\textwidth]{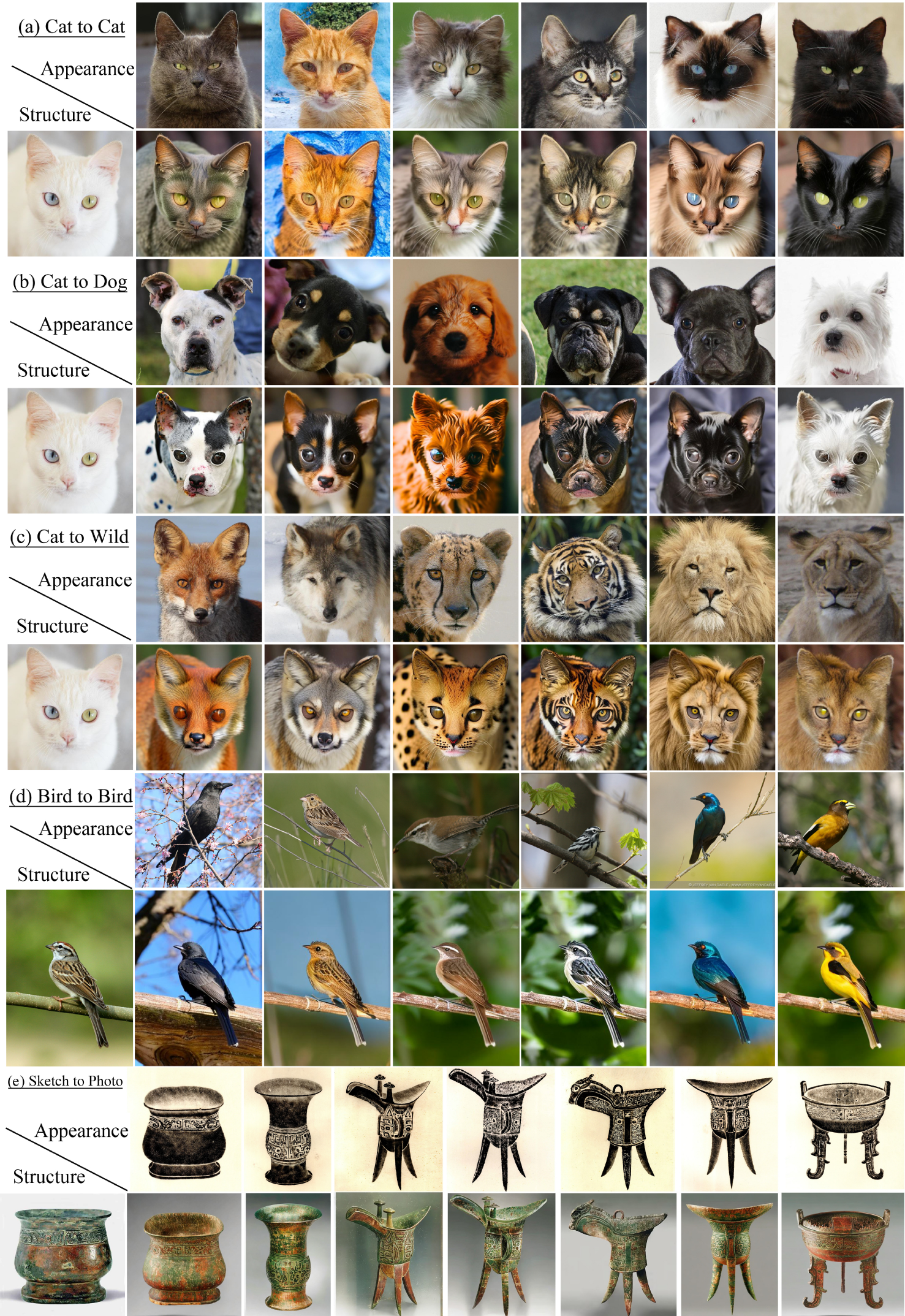}
    \caption{\small\textbf{Appearance transfer tasks.}
    We showcase the effects achieved by the baseline architecture with filtering combined operation in appearance transfer tasks.
    }
    \label{fig:5_2}
    \vspace{-6mm}
\end{figure*}
\subsection{Quantitative analysis}
Our quantitative analysis covers six specific appearance transfer: cat to cat, cat to dog, cat to wild, bird to bird, airplane to bird, and car to car. 
Following previous works~\cite{saito2020coco, liu2020unsupervised, gao2022sketchsampler, Cross-imageAttention}, we selected six metrics for our experiment.
To evaluate the retention of geometric and semantic features from structure images in the generated images, we use three primary indicators: Structure Preservation (SP), Chamfer Distance (CD), and Fréchet Inception Distance (FID). Additionally, to assess the fidelity of low-level features between appearance images and generated images, we employ three specific metrics: Gray-Level Co-occurrence Matrix (GLCM), Peak Signal-to-Noise Ratio (PSNR), and Color Histogram Correlation (CHC). 
\begin{itemize}
    \item \textit{Structure Preservation (SP)}: we utilize the marquee interaction mode of SAM~\cite{kirillov2023segment} for selecting areas to obtain binary masks corresponding to structure images and their respective output images. Then, we compute their Intersection over Union (IoU) results as a measure of Structure Preservation.
    \item \textit{Chamfer Distance (CD)}: we first extract the line drawings of the structure and generated images, and then filter out redundant details using the Canny operator. The high and low thresholds used by the Canny operator are set to 150 and 50, respectively. Finally, we calculate the chamfer distance between the line drawings as a measure of the gap between the sets of edge points in the two images. A smaller value indicates a higher degree of match between the shape or edge features in the structure and generated images.
    \item \textit{Fréchet Inception Distance (FID)}: we calculate the FID score between the structure image and the generated image to quantify the extent to which the two images align in terms of their structural features.
    \item \textit{Gray-Level Co-occurrence Matrix (GLCM)}: it is used to calculate the loss value of texture features between the appearance image and the generated image.
    \item \textit{Peak Signal-to-Noise Ratio (PSNR)}: it is used to measure how well the generated image preserves the low-level features of appearance.
    \item \textit{Color Histogram Correlation (CHC)}: it is used to calculate the color similarity between the generated image and the appearance image. Among them, we use mask to cover the background of the image.
\end{itemize}
Test Datasets: AFHQ~\cite{choi2020stargan}, CUB-200-2011~\cite{CUB-200-2011}, FGVC-Airaft~\cite{FGVC-Aircraft}, Stanford-Cars~\cite{StanfordCars}. 
Among them, the three domain data of cat, dog, and wild are all from the AFHQ test set. We followed the setting of AFHQ's test set, with 500 images for each category, and randomly selected 500 images from three other datasets as appearance images. For every types of appearance transfer tasks there are 2000 pairs, Therefore, the data in Table 
\ref{tab:Quantitative_structure} is based on the evaluation results of 12000 Structure-Appearance image pairs. 
We show examples of the comparison results in Figure~\ref{fig:5_1}.

\subsection{Qualitative analysis}
The qualitative analysis experiments include a total of five domains (cat, dog, wild, bird, bronze). 
In addition to the datasets used in the quantitative analysis experiments as shown in Figure \ref{fig:5_2}, additional data is: Bronze Dings~\cite{zhou2023multi}. In this task, the appearance reference and the structure image do not have semantic correspondence, and their relationships belong to different domains. 
So the focus in this task is to obtain the low-level texture features from the appearance image without semantic correspondence and then render it to the structure image.

\subsection{User Study} 
We conducted a user study with 18 questions evaluating generated results in structure preservation, rendering feature transfer effectiveness, and overall quality. Each participant was paid 0.5\$. To address the common issue of low-quality feedback caused by participants' lack of understanding of task, we had eight facilitators provide detailed background information to participants on campus. Participants were then asked to select the most fitting options from anonymous choices based on their preferences. Finally, we received 215 valid survey submissions, with ours garnering a support rate of 51. 89\% (cross-image 24. 00\%, IP-Adapter 14. 98\%, ControlNet + IP-Adapter 9. 13\%).
\section{Limitation and Conclusion}
We constructed an experimental framework based on IP-Adapter to explore image generation techniques. However, we found that the identity consistency of the generated results is not always satisfactory when the input image prompt contains rich semantic information. This limitation is partly because IP-Adapter may not fully balance these properties when processing different image attributes, thus affecting the consistency of the final output, which may be attempted by using mask technology to optimize identity consistency and image quality. Beyond the baseline framework utilized in this study, the integration of other advanced diffusion models may potentially lead to even superior results if they are implemented to replace certain components in the baseline framework. We will explore this further in future work.

In conclusion, we propose a pixel-space processing, FilterPrompt,  to guide image appearance transfer, by focusing on the input sensitivity and dynamic evolution of diffusion models. We find that the models adapt to input feature distributions, enabling targeted operations for precise control in final generated images. Experimental results show that our approach helps refine structural control and reduce redundant textures in transfer tasks. Although FilterPrompt requires manual setup, it provides a simple yet efficient way to enhance customization and control in diffusion models.
{
    \small
    \bibliographystyle{ieeenat_fullname}
    \bibliography{main} 
}

\end{document}